\pdfoutput=1

\documentclass[11pt]{article}

\PassOptionsToPackage{table,dvipsnames,usenames,svgnames}{xcolor}
\PassOptionsToPackage{hyphens}{url}

\usepackage{acl}

\usepackage{times}
\usepackage{latexsym}

\usepackage[T1]{fontenc}

\usepackage[utf8]{inputenc}

\usepackage{microtype}

\usepackage{inconsolata}

\usepackage{graphicx}

\usepackage{float}
\usepackage{setspace}
\usepackage{xcolor}
\usepackage{tocloft}
\usepackage{booktabs}
\usepackage{bold-extra}
\usepackage{enumitem}
\usepackage{lmodern}
\usepackage{amsfonts}
\usepackage{mathtools}
\usepackage{csquotes}
\usepackage{parskip}
\usepackage{caption}  
\usepackage{titlesec}
\usepackage{pifont}
\usepackage{hyperref}

\usepackage{mathptmx}

\newcommand{\xmark}{\ding{55}}%

\newcommand{\calc}[1]{{#1}}
\newcommand{\out}[1]{{#1}}
\newcommand{\result}[1]{{#1}}

\title{Self-training Language Models for Arithmetic Reasoning}

\author{Marek Kadlčík$^{*}$ \and Michal Štefánik$^{*}$ \vspace{5pt}\\
  Faculty of Informatics, Masaryk University, Czech Republic \vspace{5pt} \\
  \small{\texttt{\{kadlcik,stefanik.m\}@mail.muni.cz}} \vspace{0.0pt} \\
  }

\begin{document}

\maketitle

\def\thefootnote{*}\footnotetext{Equal contribution; Authors ordered alphabetically}\def\thefootnote{\arabic{footnote}}

\begin{abstract}

Recent language models achieve impressive results in tasks involving complex multistep reasoning, but scaling these capabilities further traditionally requires expensive collection of more annotated data.
In this work, we explore the potential of improving models' reasoning capabilities without new data, merely using automated feedback to the validity of their predictions in arithmetic reasoning (\textit{self-training}).

In systematic experimentation across six different arithmetic reasoning datasets, we find that models can substantially improve in both single-round (offline) and online self-training, reaching a correct result in +13.9\% and +25.9\% more cases, respectively, underlining the importance of \textit{actuality} of self-training feedback. We further find that in the single-round, \textit{offline} self-training, traditional supervised training can deliver gains comparable to preference optimization, but in \textit{online} self-training, preference optimization methods largely outperform supervised training thanks to their superior stability and robustness on unseen \textit{types} of problems.

\end{abstract}

\vspace{2em}

\section{Introduction}


Despite recent improvements in the practical usability of language models (LMs) attributed to preference alignment methods \cite{aligning_llm_human}, these models often struggle with tasks requiring \textit{reasoning}, i.e., a process of inferring a conclusion or decision logically and systematically \cite{huang-chang-2023-towards}. Previous work improves the reasoning capabilities of language models by \textit{scaling} training data to more diverse \cite{kadlcik-etal-2023-calc} or complex \cite{MATH} collections, but reaching further improvements in this direction becomes exceedingly expensive.

\vspace{0.5em}

\begin{figure}[t]
    \centering
    \includegraphics[width=0.475\textwidth]{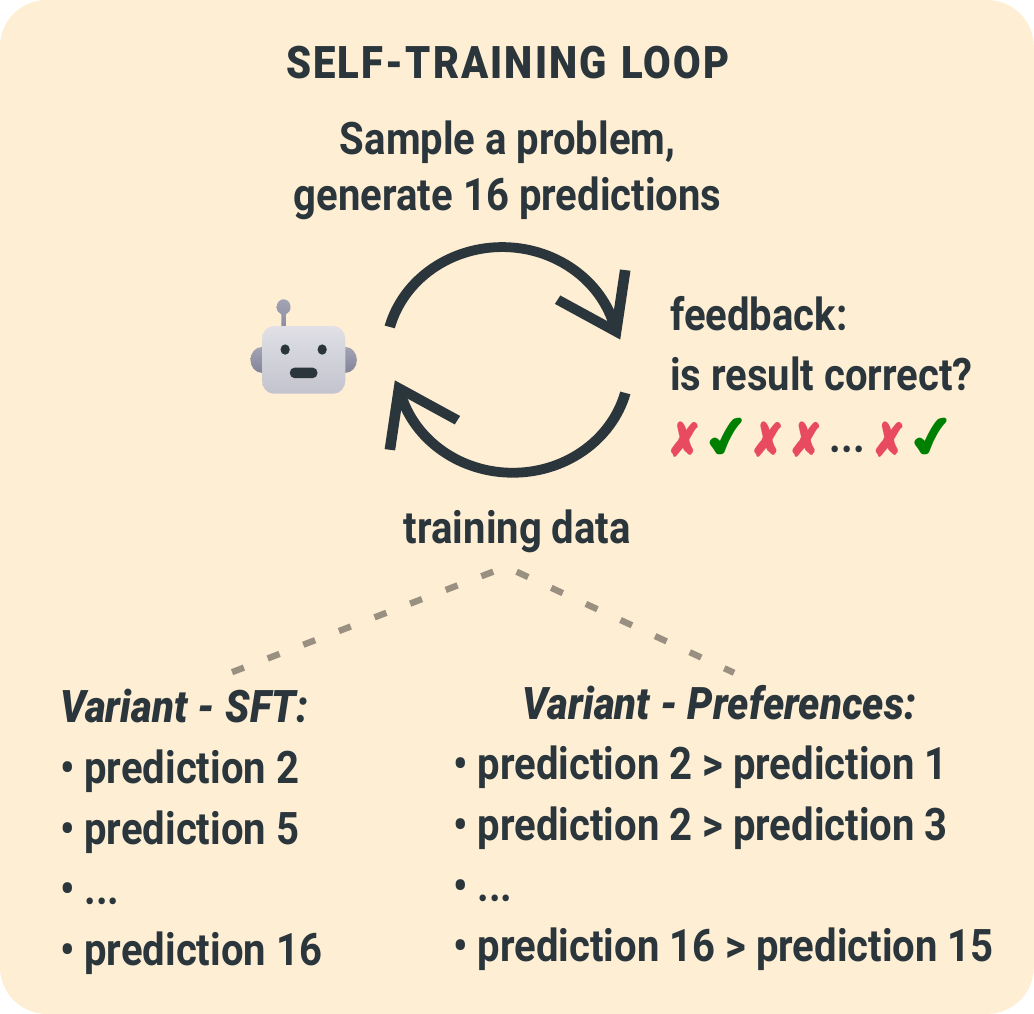}
    \caption{\vspace{-2pt}Schema of self-training that we apply to provide the model with training feedback to its predictions. In the offline variant, the model generates all predictions in a single round. In the online variant, the training data is continuously generated.
    \vspace{-5pt}
    }
    \label{figure:online-selftraining-loop}
\end{figure}

In this work, we evaluate the potential of improving models' capabilities by training from implicit, automated feedback to models' responses. \textit{Arithmetic reasoning} tasks present a challenge that reflects heavily on the model's reasoning capabilities, while the quality of the model's responses can be automatically assessed against the annotated \textit{correct results} rather than expensive and possibly subjective judgments of model outputs \cite{hu-etal-2023-decipherpref}.
Thus, we choose the arithmetic reasoning to address our two main research questions:

\noindent \textbf{RQ1}: Can we improve the reasoning abilities of language models with self-training without any new data data?

\noindent \textbf{RQ2}: Can the preference optimization bring further improvements to models' capabilities over traditional supervised fine-tuning?

We address these questions by implementing two variants of self-training: (1)~an \textit{offline} variant, where the training feedback to the model responses is constructed in a \textit{single} iteration (§\ref{section:experiment:offline-selftraining}), and (2)~an \textit{online} variant, where the model obtains and trains on the feedback to its \textit{current} predictions (§\ref{section:experiment:online-selftraining}).

Our experiments reveal that both self-training variants present an efficient method for improving LMs' capabilities with \textit{implicit} training signal; both variants allow to significantly improve the initial model without \textit{any} new data. In the offline variant, similar improvements can be achieved by both supervised and preference optimization methods. However, the online variant reveals crucial issues in scaling the supervised training to \textit{autonomous} settings. On the contrary, preference optimization methods can robustly persist the original capabilities even in autonomous self-training while reaching further improvements.

Finally, the difference in average improvement between our best-performing offline (+13.9\%) and online method (+25.9\%) indicates that the \textit{actuality} of self-training feedback is a crucial factor of self-training effectivity. 
Our results motivate future research towards exploring new sources of implicit feedback able to provide language models with immediate feedback to their current predictions.

\vspace{-4pt}
\section{Related Work}
\label{sec:related_work}
\vspace{-4pt}

We build upon a line of previous work that experiments with providing feedback to language models in arithmetical reasoning.
Notably, \citet{wizardmath} train models with PPO \citep{ppo} against feedback on individual steps given by ChatGPT 3.5. \citet{proces-vs-outcome} apply variants of self-training on GMS8K and compare the effectiveness of giving outcome-based (per solution) or process-based (per each step in solution) feedback, concluding that the two approaches result in comparable accuracy, but outcome-based feedback delivers a higher error rate in the rationales. \citet{prm800k} also focus on a comparison of process-based and outcome-based feedback on a larger scale and conclude that process-based feedback outperforms outcome-based at end-result accuracy.

Our work is closest to \citet{talm} and \citet{star}. \citet{talm} apply self-training with a traditional supervised objective: they train the model on a small set of seed data and continuously use the trained model to generate solutions for a larger set, from which correct solutions are used in another training epoch. They show that three such subsequent epochs can improve the accuracy with diminishing returns. \citet{star} experiment with self-training with supervised fine-tuning on commonsense and math reasoning. They report positive results of self-training on the model's reasoning capabilities under specific conditions: (1) the initial model must be capable enough to be able to achieve improvements, and (2) training tasks must hold a negligible chance of random success (unlike, e.g., binary classification). 

Our work builds upon these findings but differs from previous work in our objectives and data setting; We provide a systematic comparison of \textit{different} training objectives in \textit{both} online and offline settings, including the most recent preference optimization methods and show that training objective indeed plays a crucial role, especially in the online setting. Our data setting is more ambitious than of previous work: we show that self-training can deliver substantial improvements also by using \textit{only} problems already \textit{seen} in previous training.
 Finally, contrary to previous self-training work, we make our code\footnote{\url{https://github.com/prompteus/calc-x}} and models\footnote{see our \href{https://huggingface.co/collections/MU-NLPC/calcformers-selftraining-65ae7f2fbaeb4177a65bcfbc}{HuggingFace Hub}} freely available to accelerate future work in self-training.

\section{Experiments}
\label{sec:experiments}

\begin{table*}[t]
\centering
\resizebox{0.85\textwidth}{!}{
    \begin{tabular}{lcccccc}
        \toprule
        & GSM8K & AQuA-RAT & Ape210K & MAWPS & SVAMP & ASDiv-A \\
        \midrule
        \textsc{Base model}
            & 43.2±\small{2.7} 
            & 37.8±\small{6.1} 
            & 26.3±\small{2.1} 
            & 61.9±\small{4.2} 
            & 51.8±\small{3.2} 
            & 78.7±\small{2.3} 
            \\
        \midrule    
        \textsc{SFT plain}
            & \textbf{46.1}±\small{2.7} 
            & 37.8±\small{5.9} 
            & 32.9±\small{2.2} 
            & 70.6±\small{3.8} 
            & 56.2±\small{3.0} 
            & 81.9±\small{2.2} 
            \\
        \textsc{SFT plain + LoRA}
            & 44.9±\small{2.7} 
            & \textbf{39.0}±\small{5.9} 
            & \textbf{37.3}±\small{2.2} 
            & \textbf{80.8}±\small{3.5} 
            & 55.8±\small{3.1} 
            & \textbf{82.8}±\small{2.1} 
            \\
        \textsc{SFT balanced}
            & 45.8±\small{2.7} 
            & 37.4±\small{5.9} 
            & 33.6±\small{2.2} 
            & 66.7±\small{3.9} 
            & \textbf{58.4}±\small{3.0} 
            & 82.0±\small{2.2} 
            \\
        \textsc{SFT with negatives}
            & 41.8±\small{2.7} 
            & 33.1±\small{5.7} 
            & 28.0±\small{2.1}
            & 65.2±\small{4.1} 
            & 52.2±\small{3.1} 
            & 75.9±\small{2.4} 
            \\
        \midrule
        \textsc{DPO ($\beta=0.99$)}
            & 45.3±\small{2.7} 
            & 37.0±\small{5.9} 
            & 29.2±\small{2.1} 
            & 69.6±\small{3.9} 
            & 54.2±\small{3.1} 
            & 83.1±\small{2.1} 
            \\
        \textsc{DPO ($\beta=0.9$)}
            & 37.2±\small{2.6} 
            & 40.9±\small{6.1} 
            & 32.8±\small{2.3} 
            & 61.2±\small{4.1} 
            & 52.2±\small{3.1} 
            & 78.1±\small{2.3} 
            \\
        \textsc{DPO ($\beta=0.9$) + LoRA}
            & 45.9±\small{2.7} 
            & \textbf{41.3}±\small{6.1} 
            & 32.4±\small{2.2} 
            & 64.4±\small{4.0} 
            & 57.1±\small{3.1} 
            & 84.7±\small{2.0} 
            \\
        \addlinespace
        \textsc{KTO ($\beta=0.3$)}
            & \textbf{47.1}±\small{2.7} 
            & 38.6±\small{6.1} 
            & 36.4±\small{2.2} 
            & \textbf{78.3}±\small{3.5} 
            & 55.8±\small{3.1} 
            & 85.3±\small{2.0} 
            \\
        \textsc{KTO ($\beta=0.1$)}
            & 47.0±\small{2.7} 
            & 40.6±\small{6.1} 
            & \textbf{37.9}±\small{2.3} 
            & 68.3±\small{3.9} 
            & 57.2±\small{3.1} 
            & 86.4±\small{1.9} 
            \\
        \textsc{KTO ($\beta=0.1$) + LoRA}
            & 43.1±\small{2.7} 
            & 36.2±\small{5.9} 
            & 37.6±\small{2.2} 
            & 64.2±\small{4.1} 
            & 58.5±\small{3.3} 
            & 87.0±\small{1.9} 
            \\
        \addlinespace
        \textsc{IPO ($\tau=0.9$)}
            & 38.4±\small{2.7} 
            & 39.0±\small{5.9} 
            & 26.9±\small{2.1} 
            & 71.3±\small{3.8} 
            & 64.6±\small{3.0} 
            & 87.4±\small{1.9} 
            \\
        \textsc{IPO ($\tau=0.99$)}
            & 40.7±\small{2.7} 
            & 36.6±\small{5.9} 
            & 28.1±\small{2.1} 
            & 66.3±\small{4.0} 
            & 64.5±\small{3.0} 
            & \textbf{87.8}±\small{1.8} 
            \\
        \textsc{IPO ($\tau=0.99$) + LoRA}
            & 36.0±\small{2.6} 
            & 39.4±\small{5.9} 
            & 30.2±\small{2.1} 
            & 66.7±\small{4.0} 
            & \textbf{65.6}±\small{3.0} 
            & \textbf{87.8}±\small{1.8} 
            \\
        \bottomrule
        \end{tabular}
    }
    \caption{Percentage of correct results obtained in \underline{offline} self-training of \textsc{Base Model} (\textsc{Calcformer-Flan-XL}) on \textit{Ape210K} problems. For each preference optimization method, we report results for its two best-performing configurations. \textbf{Bold} entries denote the best results among supervised and preference optimization methods per dataset. Confidence intervals are bootstrapped (500 samples, 1,000 repeats).
    }
    \label{table:offline-selftraining}
\end{table*}

Our experiments build upon the 3-billion-parameter FLAN models fine-tuned specifically for arithmetic reasoning in previous work of \citet{kadlcik-etal-2023-calc}. These relatively compact calculator-assisted models called \textsc{Calcformers} were shown to perform noticeably well on multi-step reasoning, while even on single-step and two-step problems perform compared to 70B Llama-2 \cite{Touvron2023Llama2O}. Another desiderata of these models is the \textit{transparency} of their training data. In our experiments, this allows us to opt for a more challenging yet realistic self-training setting where we do \textit{not} train the models on \textit{any} \textit{new} data, but only on the problems that \textsc{Calcformers} have \textit{already} seen in the training, merely with a complementary training signal.

Specifically, we self-train these models with the prompts from Ape210K \citep{ape210k}, to our knowledge the largest available arithmetical reasoning dataset of over 200,000 math problems. In addition to Ape210K's test set, we evaluate our models on five other math datasets, assessing the robustness of models' capabilities in new \textit{types} of math problems; GSM8K \cite{gsm8k} containing multistep elementary-grade problems requiring on average 3.25 steps to achieve correct result, AQuA-RAT \cite{aqua} with more complex, multiple-choice tasks, and three simpler, one to two-steps datasets: MAWPS \cite{mawps}, ASDiv-A \cite{asdiv}, and SVAMP \cite{svamp}.

In both self-training variants, we use the trained model to generate training data (see Fig.~\ref{figure:online-selftraining-loop}). The generated data consists of the original input prompt ($x_i$) and associated model predictions ($y_i$) in the form of a chain-of-thought sequence containing the model's final result at the end. For each prompt, we generate 16 predictions using sampled generation. Annotations of correct results then allow us to automatically annotate each prediction for either being correct ($y^\textit{OK}_i$), or incorrect ($y^\textit{NOK}_i$), assigning a set of both correct and incorrect predictions to each input prompt.

For the supervised fine-tuning (SFT) objective, we construct the training dataset from pairs of ($x_i$, $y^\textit{OK}_i$). SFT uses a standard next-token prediction with cross-entropy loss and teacher forcing \cite{bahdanau2014neural}.
All preference optimization (\textsc{PO}) objectives then train on triples ($x_i$, $y^\textit{OK}_i$, $y^\textit{NOK}_i$), with the $y^\textit{OK}_i$ marked as being preferred over $y^\textit{NOK}_i$. We experiment with three recent preference optimization methods: Direct Preference Optimization; DPO \cite{dpo}, Kahneman-Tversky Optimization; KTO \cite{kto} and Identity Preference Optimization; IPO \cite{ipo}. These methods differ in a variety of aspects in the formulation of training loss. For brevity, we direct the reader to the referenced work for further details on preference optimisation methods. Further details of our general training setup can be found in Appendix~\ref{appx:training_details}.

\subsection{Offline Self-training}
\label{section:experiment:offline-selftraining}

In the offline variant, we perform a single iteration of collecting predictions with prompts from Ape210K, resulting in over 24,000 prompts with at least one positive and one negative prediction.

All \textsc{PO} methods rely on a crucial parameter $\beta$ or $\tau$ that weights the KL regularization of the trained model according to the original \enquote{reference} model. We perform a hyperparameter tuning of this parameter with $\beta \in (0.01, 0.1, 0.3, 0.6, 0.9, 0.99)$ according to in-domain validation accuracy separately for each method and report the results for the best two configurations.

For \textsc{SFT}, we experiment with 3 variants. \textsc{SFT plain} is trained on pairs ($x_i$, $y^\textit{OK}_i$). In \textsc{SFT balanced} and \textsc{SFT with negatives}, we aim to \textit{compensate} for the potential data disadvantage of \textsc{SFT plain} compared to \textsc{PO} methods exhibiting the trained model to \textit{two} solutions ($y^\textit{OK}_i$, $y^\textit{NOK}_i$) per problem:
(i)~In \textsc{SFT balanced}, we use \textit{two different} correct predictions $y^\textit{OK}_i$ for one $x_i$. 
(ii)~In \textsc{SFT with negatives}, we use both positive $y^\textit{OK}_i$ and negative $y^\textit{NOK}_i$ as targets for each $x_i$. In the training data constructed from $y^\textit{NOK}_i$, we prefix $x_i$ with a phrase \enquote{Write incorrect solution for the following problem}. This exposes the model to both correct and incorrect solutions, conceivably helping it to differentiate between the two within SFT training. 


Finally, we re-train the best-performing run of each method with a low-rank adaptation (LoRA)~\citep{lora}, a commonly used fine-tuning regularization technique that restricts the fine-tuning update of each weight to have a specific low rank. We apply LoRA with a rank of 32 on all linear projections in the model.

\begin{table*}[t]
    \centering
    \resizebox{0.85\textwidth}{!}{
        \begin{tabular}{lcccccc}
            \toprule
            & GSM8K & AQuA-RAT & Ape210K & MAWPS & SVAMP & ASDiv-A \\
            \midrule
            \textsc{Toolformer} (6.7B)
                & 
                & 
                & 
                & 44.0 
                & 29.4 
                & 40.4 
                \\
            \textsc{Llama }2 (70B)
                & 
                & 
                & 
                & 82.4 
                & 69.2 
                & 67.1 
                \\
            \midrule
            \textsc{Base model (3B)}
                & 43.2±\small{2.7} 
                & 37.8±\small{6.1} 
                & 26.3±\small{2.1} 
                & 61.9±\small{4.2} 
                & 51.8±\small{3.2} 
                & 78.7±\small{2.3} 
                \\
            \midrule
            \textsc{SFT}
                & 27.4±\small{2.5} 
                & 7.9±\small{3.3} 
                & 41.2±\small{2.3} 
                & 63.8±\small{4.2} 
                & 59.8±\small{3.1} 
                & 83.3±\small{2.1} 
            \\
            \textsc{DPO ($\beta=0.9$)}
                & 49.1±\small{2.7} 
                & \textbf{39.8}±\small{5.9} 
                & 37.9±\small{2.3} 
                & 79.6±\small{3.4} 
                & 57.3±\small{3.1} 
                & 85.6±\small{2.0} 
            \\
            \textsc{KTO ($\beta=0.1$)}
                & \textbf{52.7}±\small{2.7} 
                & 36.6±\small{6.1} 
                & \textbf{49.6}±\small{2.4} 
                & \textbf{85.2}±\small{3.0} 
                & \textbf{62.6}±\small{3.1} 
                & \textbf{90.6}±\small{1.6} 
            \\
            \textsc{IPO ($\tau=0.99$)}
                & 49.1±\small{2.8} 
                & 35.8±\small{5.9} 
                & 42.2±\small{2.3} 
                & 81.5±\small{3.4} 
                & 56.8±\small{3.0} 
                & 86.6±\small{1.9} 
            \\
            \bottomrule
        \end{tabular}
    }
    \caption{
        Percentage of correct results obtained by \underline{online} self-training of \textsc{Base Model} (\textsc{Calcformer-Flan-XL}) on Ape210K problems. \textbf{Bold} denotes the best self-trained result per dataset.
        Confidence intervals are obtained from bootstrapping (500 samples, 1,000 repeats).
        Evaluations of the previous tool-using arithmetic reasoning models (\textsc{Toolformer} and \textsc{Llama 2}) are self-reported results from \citet{toolformer} and \citet{Touvron2023Llama2O}, and are limited to single-step reasoning datasets due to inherent limitations of their tool-using mechanism.
    }
    \label{table:online-selftraining}
\end{table*}

\paragraph{Results}
Table~\ref{table:offline-selftraining} compares the accuracy achieved in offline self-training with each method.
A comparison of supervised and more complex preference optimization methods reveals a relatively small difference between the best-performing configurations of both categories. Especially thanks to LoRA regularization, \textsc{SFT} shows the ability to reach results comparable in most datasets. 
Similar to \textsc{SFT}, LoRA regularization also has a positive effect on DPO, evidencing DPO's inclination to overfitting, as also evidenced by previous work~\citep{ipo}.
Among all supervised methods, the \textsc{SFT with negatives} performs the worst, showing that using negative feedback in supervised training analogically to preference optimization is non-trivial.

On the practical side, we note that \textsc{PO} methods converge much faster than \textit{SFT} methods, achieving the best validation scores on average after around 2,400 training steps compared to 16,600 steps in supervised setups.
A detailed comparison of training steps and time can be found in Table~\ref{table:offline-selftraining-speed}.



\subsection{Online Self-training}
\label{section:experiment:online-selftraining}

In the online self-training, we generate the training data on the fly. Therefore, throughout the whole training, both the positive and negative predictions used for conditioning the updates can realistically be generated by the trained model. Previous work showed that exposing the model to its own outputs might itself improve its robustness \cite{stefanik-etal-2023-soft}. In our online self-training experiments, we additionally evaluate the LM's capability to autonomously improve its \textit{reasoning} capability based on the up-to-date feedback to its own predictions.

A methodology of constructing training samples from the model's predictions for both \textsc{SFT} and \textsc{PO} methods remains identical to the offline variant. Details of data processing can be found in Appendix~\ref{appx:training_details:online}. As the generation process in online training substantially slows down updates, we restrain the scale of experiments to the best-performing configurations from the offline variant.

\paragraph{Results}
Table~\ref{table:online-selftraining} shows the accuracy of training methods in online self-training. 
This setting reveals much larger differences between methods. Supervised fine-tuning (\textsc{SFT}) improves accuracy on simple one-step and two-step datasets (MAWPS, SVAMP, and ASDiv-A) but substantially degrades performance on out-of-distribution GSM8K and AQuA-RAT. Manual inspection (Appendix~\ref{appx:analysis}) reveals that the degradation on AQuA-RAT is caused by the model's forgetting of the response format of multiple-choice questions, well-preserved by all \textsc{PO} methods.

Contrary to the \textsc{SFT}, PO methods deliver significant improvements compared to \textit{both} the base model \textit{and} their offline variants (Table~\ref{table:offline-selftraining}). Noticeable is the improvement of DPO on GSM8K (by 11.9\% of absolute accuracy, i.e. by 22.0\% relative to base model), among other cases, suggesting that self-training can mitigate overfitting of PO methods. The best-performing KTO method also substantially improved compared to the offline variant; by 11.3\% of accuracy on in-domain Ape210K, or by 16.9\% on simpler, out-of-domain MAWPS.
Among all other online methods, KTO performs best on every dataset except for AQuA-RAT, on average improving by 12.9\% of absolute accuracy, i.e. by 25.9\% relative to the base model.

Appendix~\ref{appx:analysis} provides a per-sample analysis of differences between outputs of SFT and PO models, with a report from a manual assessment of faithfulness of models' rationales in Table~\ref{table:errors}. Noticeably, we find that while the SFT also achieves large in-distribution improvements, this comes for the price of faithfulness and usability of its rationales, as the SFT model learns to completely or partially omit most of the rationales.

Figure~\ref{fig:kto-progress} visualizes the dynamics of online self-training in solving known problems during training. We can see that self-training increases the proportion of problems that it always solves correctly and, more importantly, robustly reduces the proportion of problems that it can \textit{not} solve.

\section{Conclusions}
\label{sec:conclusion}

This work explores the potential of autonomously improving language models for arithmetic reasoning: a task allowing automated, immediate, and objective feedback based on the correct results. We experiment with two settings: (i)~\textit{offline self-training}, collecting the feedback in a single iteration, and (ii)~\textit{online self-training}, where the model trains continuously from feedback to its up-to-date predictions. In both settings, we apply and compare recent preference optimization methods (DPO, KTO, IPO) with standard supervised training (SFT).

We find that self-training provides an opportunity to improve models' capabilities without \textit{any} new data, using \textit{exclusively} models' own predictions and automated feedback. In addition to the offline variant, online self-training provides further opportunities for data-free improvements thanks to the enhanced robustness of preference optimization methods. 

Our work motivates future work towards seeking other sources of implicit training feedback beyond arithmetic reasoning, exemplified in previous work in a reasoning coherence \cite{akyurek-etal-2024-deductive} or consistency \cite{stefanik2023conceptaware}. Presenting language models with novel sources of implicit feedback via self-training can fill the gap of the traditional, largely simplified training objectives and empower models to capture more complex structural dependencies necessary in many real-world applications.

\section*{Limitations}

Despite the fact that our proposed self-training methods do not require any new human annotation, we acknowledge their limitations in the extensive computational requirements given by generating the data. While the data generation for the offline variant can be parallelized, this is more difficult for the online variant, where the model is trained with its own most recent predictions. As a result, our self-training experiments took between 15 and 30 days to converge on a single Nvidia A100 GPU. 

The time-demanding character of online self-training experiments is a direct cause of another limitation: a constrained diversity of models and datasets that we experiment with. As such, the experiments and conclusions of our work should inspire experiments with self-training in other applications but may not be generalized to claims on the general effectiveness of self-training.

\section*{Acknowledgements}

We acknowledge the Centre for Biomedical Image Analysis at Masaryk University supported by MEYS CR (LM2023050 and CZ.02.1.01/0.0/0.0/18\_046/0016045 Czech-BioImaging) for providing computational resources for training models and collecting evaluations presented in this paper.

\bibliography{bibliography}

\appendix

\section{Training Details}
\label{appx:training_details}

In every configuration of both preference and supervised training, the model is trained with Adafactor~\citep{adafactor} optimizer with an effective batch size of 32, a~learning rate of $2\cdot10^{-5}$ with 1,000 warmup steps, and a linear decay to~0 in~1~million steps. The models were trained in bfloat16~\citep{bfloat16} precision with mixed precision training~\citep{mixed-precision-training}. The training terminates after convergence on the in-domain dataset (Ape210K), and then the best checkpoint from the training is selected according to in-domain validations.

Each of our experiments can be reproduced with a single Nvidia A100/A40 graphic card and 32GB of RAM. Note that especially the online self training experiments can take up to 31 days to converge.

\begin{table}[tbh]
    \centering
    \resizebox{1.0\columnwidth}{!}{
        \begin{tabular}{lrr}
            \midrule
            Method & Training steps & Wall Time \\
            \midrule
            \textsc{SFT plain} & 16,000 & 17 h \\
            \textsc{SFT plain + LoRA} & 98,000 & 120 h \\
            \textsc{SFT balanced} & 14,000 & 15 h \\
            \textsc{SFT with negatives} & 20,000 & 21 h \\
            \midrule
            \textsc{DPO $\beta=0.99$} & 1,800 & 2 h \\
            \textsc{DPO $\beta=0.9$} & 1,800 & 2 h \\
            \textsc{DPO $\beta=0.9$ LoRA} & 2,600 & 6 h \\
            \addlinespace
            \textsc{KTO $\beta=0.3$} & 3,800 & 7 h \\
            \textsc{KTO $\beta=0.1$} & 4,800 & 8 h \\
            \textsc{KTO $\beta=0.1$ LoRA} & 16,400 & 35 h \\
            \addlinespace
            \textsc{IPO $\tau=0.9$} & 1,200 & 2 h \\
            \textsc{IPO $\tau=0.99$} & 1,200 & 2 h \\
            \textsc{IPO $\tau=0.99$ LoRA} & 1,600 & 4 h \\
            \midrule
        \end{tabular}
    }

    \caption{Number of steps and wall time that different methods take until convergence in \underline{offline} self-training shows that preference optimization methods converge 5--20 times faster than supervised training. Note that wall time fluctuates based on hardware usage by other programs and should be taken as an approximate measure.
    }
    \label{table:offline-selftraining-speed}
\end{table}

\subsection{Online self-training}
\label{appx:training_details:online}

To create new data in online self-training, we sample a random problem from Ape210K and generate predictions with the current model. Next, we label each solution as correct if its result matches the one in the data. The online self-training process is illustrated in Figure~\ref{figure:online-selftraining-loop}.

In this experiment, we again compare supervised training and preference optimization. In all variants, we generate 16 solutions per problem with top-k=50 sampling using the latest model, but the subsequent data processing is method-specific.

\paragraph{Supervised training:} After generating the solutions, we discard the incorrect ones. The correct solutions are oversampled to generate 32 training examples. Each solution is sampled at most 4 times each, and all solutions are used almost the same number of times (maximal difference of one).

\paragraph{Preference Optimization:} After the solutions are generated, we create all possible pairs of solutions where one solution has a correct result and the other one does not. We then sample with repetition from the pairs, such that:
\begin{enumerate}
    \item \label{item:po-data-oversampling-condition} every correct solution is used at most 4 times,
    \item the number of preference pairs per problem is 32 if possible without violating the condition~\ref{item:po-data-oversampling-condition},
    \item all correct solutions are used almost the same number of times,
    \item all incorrect solutions are used almost the same number of times.
\end{enumerate}
Almost the same number of times means a maximal difference of one.

In both supervised and preference training, the training instances are put into a buffer with 8192 slots, from which they are sampled randomly for training. When a batch of data gets sampled, it is removed from the buffer, and new data are generated with the correct model to fill the empty slots.

\begin{figure}[th]
    \centering
    \includegraphics[width=0.975\linewidth]{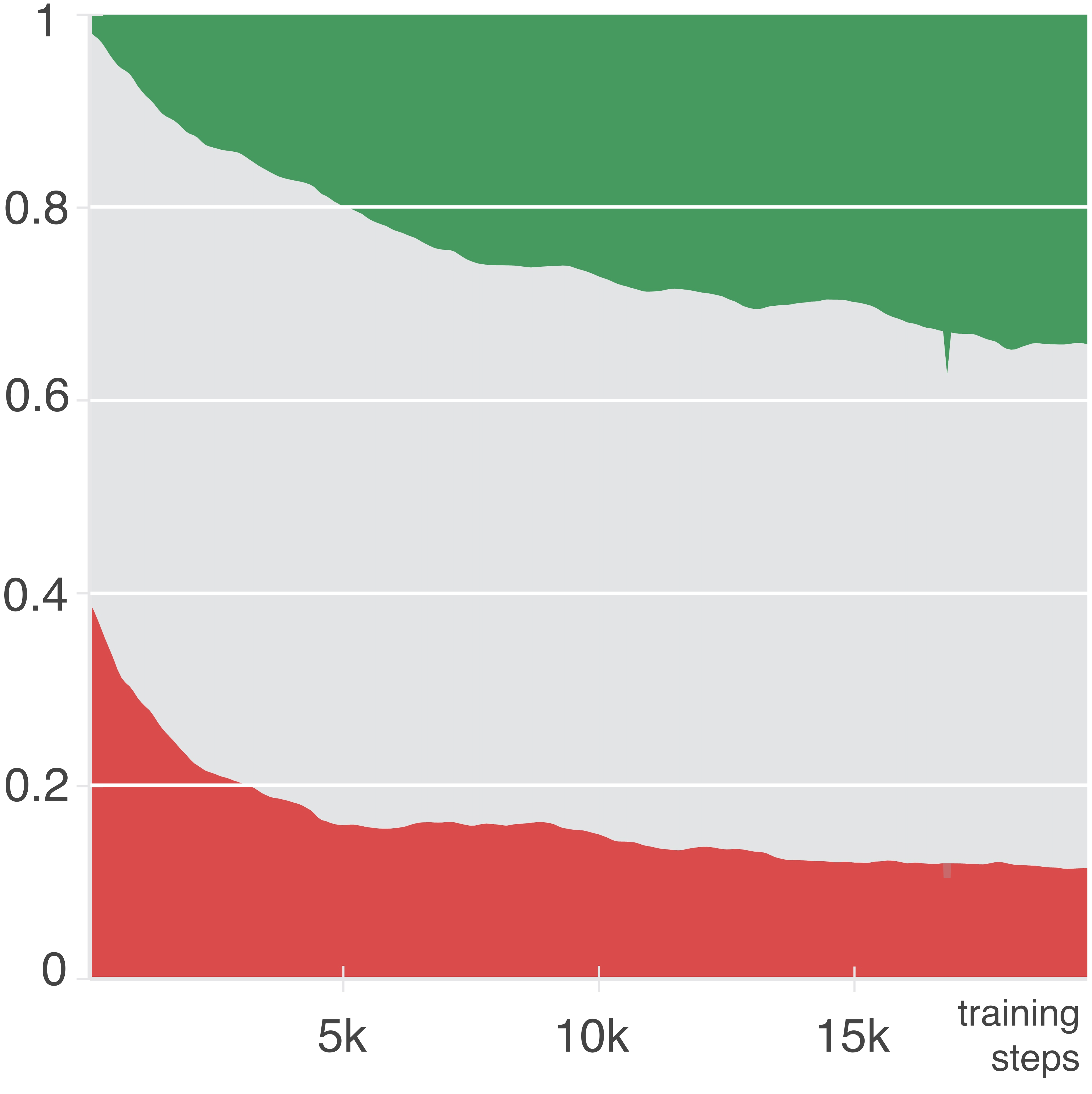}
    \caption{\textbf{Training dynamics of online training:} The fraction of training problems for which the model predicted all \textcolor{ForestGreen}{\rule{2mm}{2mm}} and none \textcolor{red}{\rule{2mm}{2mm}} of 16 trials correctly during training of the online KTO with $\beta=0.1$. The fraction is computed from a sliding window of the last 1000 problems and the chart is smoothed for visual clarity.}
    \label{fig:kto-progress}
\end{figure}

During training, we track the proportion of problems that the models consistently solve correctly or fail to solve across 16 trials. Figure~\ref{fig:kto-progress} shows the progression of the best-performing \textit{online} training run elaborating the preference optimisation with KTO.

\section{Output analyses}
\label{appx:analysis}
Aiming to better understand the difference between self-training with preference optimization methods and supervised training, we manually analyze a set of randomly chosen rationales generated for prompts of the GSM8K test set. We collect the rationales from (i) the original checkpoint, (ii) the checkpoint trained in online self-training and supervised method (denoted SFT), and (iii) the checkpoint trained on online self-training with the best-performing method (KTO). Due to the time complexity of evaluating long chain-of-thought output sequences, we analyze 20 predictions marked as correct for each checkpoint.

Within the analysis, we encounter 5 types of dominant flaws that models' outcomes exhibit, even when being correct:
\begin{enumerate}
    \item \textbf{Inconsistency:} Within the rationale, the model generates a new reasoning step which is not logically consistent with previous ones.
    \item \textbf{Missing association:} Model's rationale contains steps that are difficult to assess for consistency, as they lack the associations of units (e.g., of size, distance, or volume) or subjects from input prompt or intermediate computation.
    \item \textbf{Missing rationale:} Model only generates the result without any rationale associated with it.
    \item \textbf{Missing rationale part:} Model's rationale is missing a specific segment, making it impossible to fully check the model's computation process.
    \item \textbf{Not understandable:} Model's rationale contains text that is incomprehensible by the annotator, thus impossible to judge for logical correctness.
\end{enumerate}

The results of this analysis are summarized in Table~\ref{table:errors}. A set of predictions for identical prompts and responses of SFT and KTO checkpoints can also be found in Appendix~\ref{appx:examples}.

\begin{table}[h]
    \centering
    \resizebox{\columnwidth}{!}{
        \begin{tabular}{llll}
        \hline
                               & Original & SFT   & KTO  \\ \midrule
        Inconsistency          & 20\%     & 5\%   & 30\% \\ 
        Missing association    & 0\%      & 70\%  & 0\%  \\ 
        Missing rationale      & 0\%      & 30\%  & 0\%  \\ 
        Missing ratio. part & 5\%      & 100\% & 15\% \\ 
        Not understandable     & 30\%     & 0\%   & 30\% \\ \hline
        \end{tabular}
    }

    \caption{Output analysis: ratio of model predictions containing one of our identified flaws, evaluated on a sample of models' \textit{correct} predictions.}
    \label{table:errors}
\end{table}

\subsection{Examples of predictions}
\label{appx:examples}

For an illustration of differences between the supervised and preference optimization checkpoints, below we include a set of predictions of SFT and KTO checkpoints for randomly selected prompts of GSM8K collection. These examples can also be found by associated IDs in the original dataset repository\footnote{\url{https://huggingface.co/datasets/MU-NLPC/Calc-gsm8k}}. For better readability, we removed the navigation tokens used by the model for interaction with the calculator tool during the generation.

{
\parindent0pt 
\setlength{\parskip}{0.5em}

    \vspace{1em}
    
    \paragraph{id: gsm8k\_\_aBNdImSK9ddCwuYb} \mbox{}
    
    \textit{Question:} Tobias bought a big pizza with 60 pieces. He ate 2/5 of the pieces on the first day, 10 pieces on the second day, and 7/13 of the remaining pieces on the third day. How many pizza pieces has he eaten so far?
    
    \textit{Correct answer:} 48
    
    \textit{Prediction with SFT:}
    
    \begin{quote}
    60 * 2/5 =
    \result{28} \xmark
    \end{quote}
    
    \textit{Prediction with KTO:}
    
    \begin{quote}
    On the first day, Tobias ate 60 * 2/5 =
    \calc{60*2/5}
    \out{24}
    24 pieces. The number of pieces that remained after the second day is 60 - 24 =
    \calc{60-24}
    \out{36}
    36 pieces. On the third day, he ate 36 * 7/13 =
    \calc{36*7/13}
    \out{252/13 = around 19.384615}
    25 pieces. The total number of pieces he has eaten so far is 10 + 252/13 =
    \calc{10+252/13}
    \out{382/13 = around 29.384615}
    38 pieces. Tobias ate a total of 24 + 10 + (382/13) =
    \calc{24+10+352/13}
    \out{794/13 = around 61.076923}
    69 pieces.
    \result{794/13 = around 69.076923} \xmark
    \end{quote}

    \vspace{1em}
    
    \paragraph{id: gsm8k\_\_qIygQI2WNfRPWoYj} \mbox{}
    
    \textit{Question:} Jennie is helping at her mom's office. She has a pile of 60 letters needing stamps, and a pile of letters already stamped. She puts stamps on one-third of the letters needing stamps. If there are now 30 letters in the pile of already-stamped letters, how many were in that pile when Jennie began?
    
    \textit{Correct answer:} 10
    
    \textit{Prediction with SFT:}
    
    \begin{quote}
    60 / 3 =
    \result{40} \xmark
    \end{quote}

    \textit{Prediction with KTO:}
    
    \begin{quote}
    Jennie stamps 60 / 3 =
    \calc{60/3}
    \out{20}
    20 letters. Then the number of letters still needing stamps is 60 - 20 =
    \calc{60-20}
    \out{40}
    40. The number of letters already stamped is 30 - 20 =
    \calc{30-20}
    \out{10}
    10. There were 40 + 10 =
    \calc{40+10}
    \out{50}
    50 letters in the pile when Jennie began.
    \result{50} \xmark
    \end{quote}

    \vspace{1em}
    
    \paragraph{id: gsm8k\_\_0oOjz5Ub66DF4inZ} \mbox{}
    
    \textit{Question:} There are 6 trees in Chris's yard. Ferdinand has half the number of trees that Chris has. Harry has 5 more than twice the number of trees that Ferdinand has. How many more trees are in Harry's yard than Ferdinand's yard?
    
    \textit{Correct answer:} 8

    \textit{Prediction with SFT:}
    
    \begin{quote}
    \result{9} \xmark
    \end{quote}

    \textit{Prediction with KTO:}
    
    \begin{quote}
    Half the number of trees that Chris has is 6 trees / 2 =
    \calc{6/2}
    \out{3}
    3 trees. Twice the number of trees that Ferdinand has is 3 trees * 2 =
    \calc{3*2}
    \out{6}
    6 trees. Harry has 5 trees + 6 trees =
    \calc{6+5}
    \out{11}
    11 trees. There are 11 trees - 6 trees =
    \calc{11-6}
    \out{5}
    5 more trees in Harry's yard than in Ferdinand's yard.
    \calc{5}
    \out{5}
    5 more trees in Harry's yard than in Ferdinand's yard.
    \calc{5}
    \out{5}
    5 more trees in Harry's yard than in Ferdinand's yard.
    \calc{11}
    \out{11}
    5 more trees than in Ferdinand's yard.
    \calc{11-6}
    \out{5}
    5 more trees in Harry's yard.
    \result{5} \xmark
    \end{quote}
    
}

\end{document}